\documentclass[10pt,twocolumn,letterpaper]{article}

\usepackage{cvpr}              

\usepackage{graphicx}
\usepackage{amsmath}
\usepackage{amssymb}
\usepackage{booktabs}
\usepackage{bm}
\usepackage{multirow}
\usepackage[accsupp]{axessibility}

\usepackage[pagebackref,breaklinks,colorlinks]{hyperref}

\usepackage[capitalize]{cleveref}
\crefname{section}{Sec.}{Secs.}
\Crefname{section}{Section}{Sections}
\Crefname{table}{Table}{Tables}
\crefname{table}{Tab.}{Tabs.}

\def\cvprPaperID{948} 
\def\confName{CVPR}
\def\confYear{2022}

\begin{document}

\title{Attribute Group Editing for Reliable Few-shot Image Generation}

\author{
Guanqi Ding$^{1,2}$\footnotemark[1] ,
Xinzhe Han$^{1,2}$\footnotemark[1] ,
Shuhui Wang$^{2,4}$\footnotemark[2] ,
Shuzhe Wu$^{3}$,
Xin Jin$^{3}$,
Dandan Tu$^{3}$,
Qingming Huang$^{1,2,4}$\\
$^{1}$University of Chinese Academy of Sciences, Beijing, China\\
$^{2}$Key Lab of Intell. Info. Process., Inst. of Comput. Tech., CAS, Beijing, China\\ 
$^{3}$Huawei Cloud EI Innovation Lab, China ~~  $^{4}$Peng Cheng Laboratory, Shenzhen, China\\
 {\tt\small \{dingguanqi19, hanxinzhe17\}@mails.ucas.ac.cn, wangshuhui@ict.ac.cn,}\\
 {\tt\small  \{wushuzhe2, jinxin11, tudandan\}@huawei.com,  qmhuang@ucas.ac.cn}
}
\maketitle
 
\renewcommand{\thefootnote}{\fnsymbol{footnote}} 
\footnotetext[1]{These authors contributed equally to this work.} 
\footnotetext[2]{Corresponding authors.}

\begin{abstract}
   Few-shot image generation is a challenging task even using the state-of-the-art Generative Adversarial Networks (GANs). Due to the unstable GAN training process and the limited training data, the generated images are often of low quality and low diversity.
   In this work, we propose a new ``editing-based" method, {\it i.e.}, Attribute Group Editing (AGE), for few-shot image generation. The basic assumption is that any image is a collection of attributes and the editing direction for a specific attribute is shared across all categories. 
   AGE examines the internal representation learned in GANs and identifies semantically meaningful directions. Specifically, the class embedding, {\it i.e.}, the mean vector of the latent codes from a specific category, is used to represent the category-relevant attributes, and the category-irrelevant attributes are learned globally by Sparse Dictionary Learning on the difference between the sample embedding and the class embedding.
   Given a GAN well trained on seen categories, diverse images of unseen categories can be synthesized through editing category-irrelevant attributes while keeping category-relevant attributes unchanged. 
   Without re-training the GAN, AGE is capable of not only producing more realistic and diverse images for downstream visual applications with limited data but achieving controllable image editing with interpretable category-irrelevant directions. Code is available at~\href{https://github.com/UniBester/AGE}{https://github.com/UniBester/AGE}.
   
\end{abstract}








\section{Introduction}
\label{sec:intro}
\begin{figure}[t]
\centering
     \includegraphics[width=0.9\linewidth]{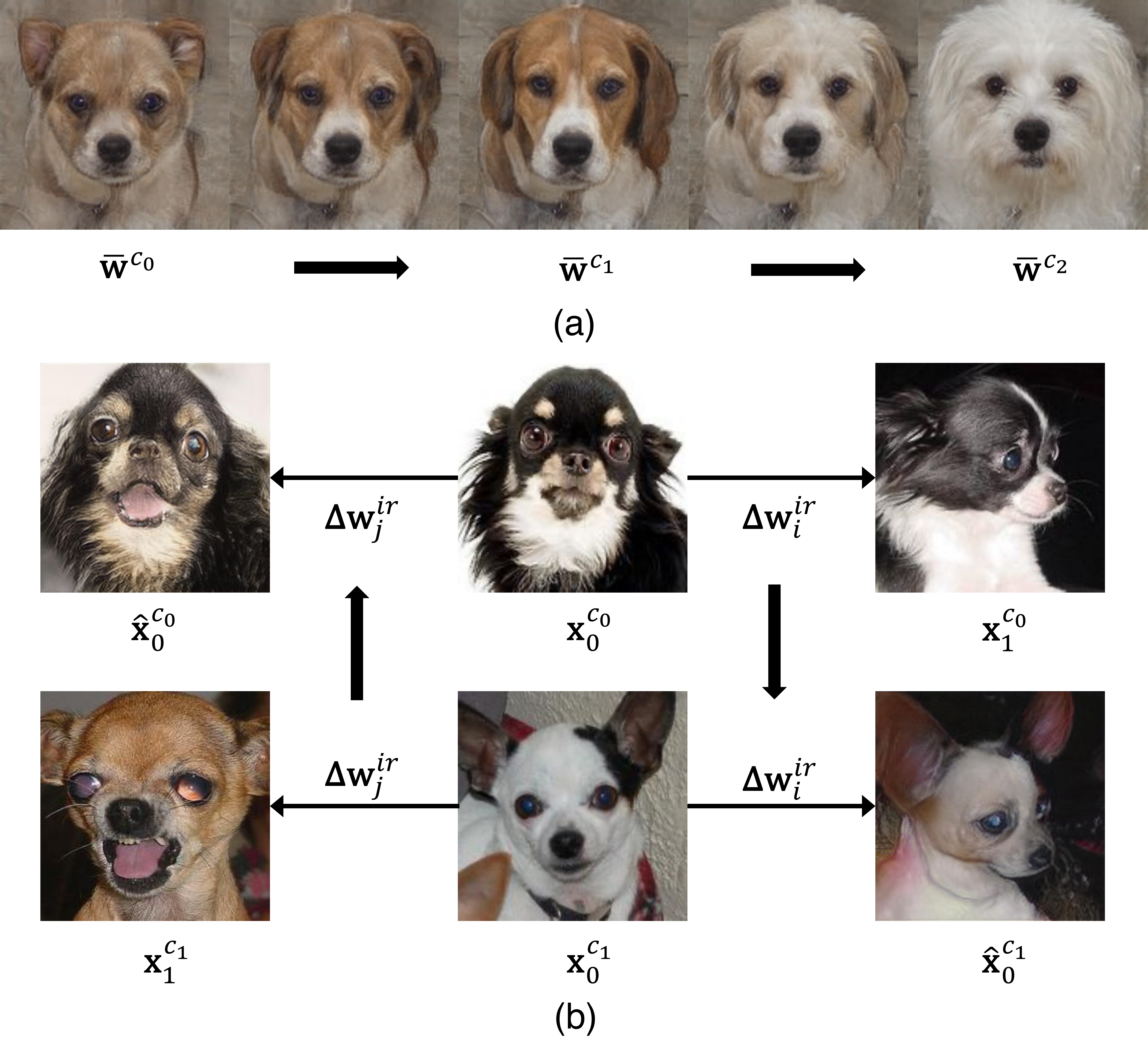}
     \vspace{-1em}
    \caption{\textbf{Illustration of attribute editing in the latent space.} 
    (a) The category-relevant attribute editing between the class embedding $\overline{\mathbf{w}}^{c_m}$. (b) The category-irrelevant attribute editing within categories. The same editing in the latent space is universal between different categories} 
    \vspace{-2em}
    \label{fig:introduction}
\end{figure}

The emergence of GANs~\cite{GAN} has enabled the deep generative model to generate images with higher quality and diversity. 
However, due to the characteristics of GANs, training a good GAN model often requires a lot of training data and is difficult to converge~\cite{DataAugGAN}. Given a few images from an unseen category, it is hard for GANs to generate new realistic and diverse images for this category. This task is referred to as few-shot image generation, which can benefit a bunch of downstream tasks like low-data detection~\cite{FewShotDetection} and few-shot classification~\cite{FewShotClassification0, FewShotClassification1}. 


Existing few-shot image generation methods can be roughly divided into three types, {\it i.e.},  optimization-based~\cite{figr, dawson}, fusion-based~\cite{gmn, f2gan, matchinggan, lofgan}, and transformation-based~\cite{dagan, deltagan}. Optimization-based methods introduce meta-learning paradigms to generate new images by learning an initialized base model and fine-tuning the model with each unseen category, but images generated by methods of this type are blurry and of low quality. Fusion-based methods fuse several input images in a feature space and decode the fused feature back to a realistic image of the same category. However, these methods need at least two images as input and can only generate images similar to input images. Transformation-based methods find intra-category transformations and apply these transformations to unseen category samples to generate more images of the same category. 
However, the end-to-end training of image transformation and generation is very unstable.
Worse still, these transformations tends to be complicated, and the generated images are often of low-quality and even crashed.

Drawing on the idea of StyleGAN~\cite{stylegan, stylegan2}, an image can be regarded as a collection of different attributes. 
The category of an image is decided by objects' category-relevant attributes, such as the shape of the face and the morphology of the fur.
Differences between images of the same category are reflected in category-irrelevant attributes including expressions, postures, \etc.
Moreover, many works~\cite{eigengan, interfacegan} have shown that GANs represent these attributes in the latent space. Moving the latent code along an identified direction can accordingly change the semantic in the output images.
Theoretically, given a pretrained GAN, an object of an unseen category can be generated with combinations of attributes from seen categories as shown in Figure~\ref{fig:introduction}(a). 
Diverse images of the same category can be generated by editing category-irrelevant attributes, which are shared across all categories as shown in Figure~\ref{fig:introduction}(b).
If these semantically meaningful directions can be distinguished, we can achieve reliable few-shot image generation needless of re-training a GAN. 

In order to identify such directions, image editing methods typically annotate a collection of synthesized samples and train linear classifiers in the latent space.
They require a clear definition of the target attributes as well as the corresponding manual annotations.
However, it will be unrealistic to obtain such detailed annotations for more complicated multi-category image generation~\cite{animalfaces, flowers, vggfaces}.
Therefore, the key challenge of the editing-based method is to factorize the meaningful directions for category-relevant and category-irrelevant attributes without explicit supervision.

To achieve this goal, we propose \textit{Attribute Group Editing (AGE)}, which examines the variation relationship between the image and the internal representation.
The core of AGE is to factorize the directions of category-irrelevant attributes and category-relevant attributes without explicit supervision.
First, the class embedding for a specific category is obtained from the mean representation of all samples from this category. 
As shown in Figure~\ref{fig:introduction}(a), for a seen category with a large amount of training data, this embedding is likely to disentangle all category-relevant attributes from major category-irrelevant attributes. 
Afterward, any samples in the dataset can be regarded as category-irrelevant editing from the corresponding class embedding.
In order to factorize category-irrelevant directions in latent space, we model this editing process with Sparse Dictionary Learning (SDL)~\cite{sdl, ksvd}.
A number of constraints are used to ensure that every direction in the dictionary is semantically meaningful and category-irrelevant. 
Different linear combinations of directions in the dictionary can facilitate generation of diverse images without changing their category.

Our contributions can be summarized as follows:

- We present a new perspective for few-shot image generation, {\it i.e.}, diverse images of unseen categories can be produced through category-irrelevant image editing.

- We propose a new method, Attribute Group Editing (AGE), which can identify groups of category-relevant and category-irrelevant editing directions from a pretrained GAN without explicit supervision. 

- Extensive experiments suggest that AGE achieves more stable few-shot image generation with high quality and diversity. Besides, since the editing directions discovered by AGE are semantically meaningful, we can also perform controllable image generation based on the learned attribute dictionary.

\section{Related Work}

\noindent\textbf{Few-shot image generation.} Existing few-shot image generation methods can be roughly divided into optimization-based methods, fusion-based methods, and transformation-based methods. Optimization-based methods~\cite{figr, dawson, metafew} combine meta-learning and adversarial learning to generate images of unseen category by fine-tuning the model. However, the images generated by such methods have poor authenticity. Fusion-based methods fuse the features by matching the random vector with the conditional images~\cite{matchinggan} or interpolate high-level features of conditional images 
by filling in low-level details~\cite{f2gan,lofgan}.
Simple content fusion limits the diversity of generated images. Transformation-based methods \cite{dagan, deltagan} capture the cross-category or intra-category transformations to generate novel data of unseen categories. These works capture the transformations from the image differences and may corrupt due to the complex transformations between intra- and inter-category pairs. 
From our new ``editing-based" perspective, the intra-category transformation can be alternatively modeled as category-irrelevant image editing based on one sample instead of pairs of samples. 

\noindent\textbf{Few-shot image-to-image translation.} Few shot image-to-image translation methods map images from one domain to another based on a few images, like category transfer~\cite{animalfaces,munit,snit}, weather transfer~\cite{manifest} and style transfer~\cite{cdimage,ewc,cdone}. 
These methods also focus on the few-shot setting but mainly handles domain transfer rather than object categories.

\begin{figure*}[ht]
    \centering
    \includegraphics[width=0.95\linewidth, trim=2cm 0em 2cm 0em]{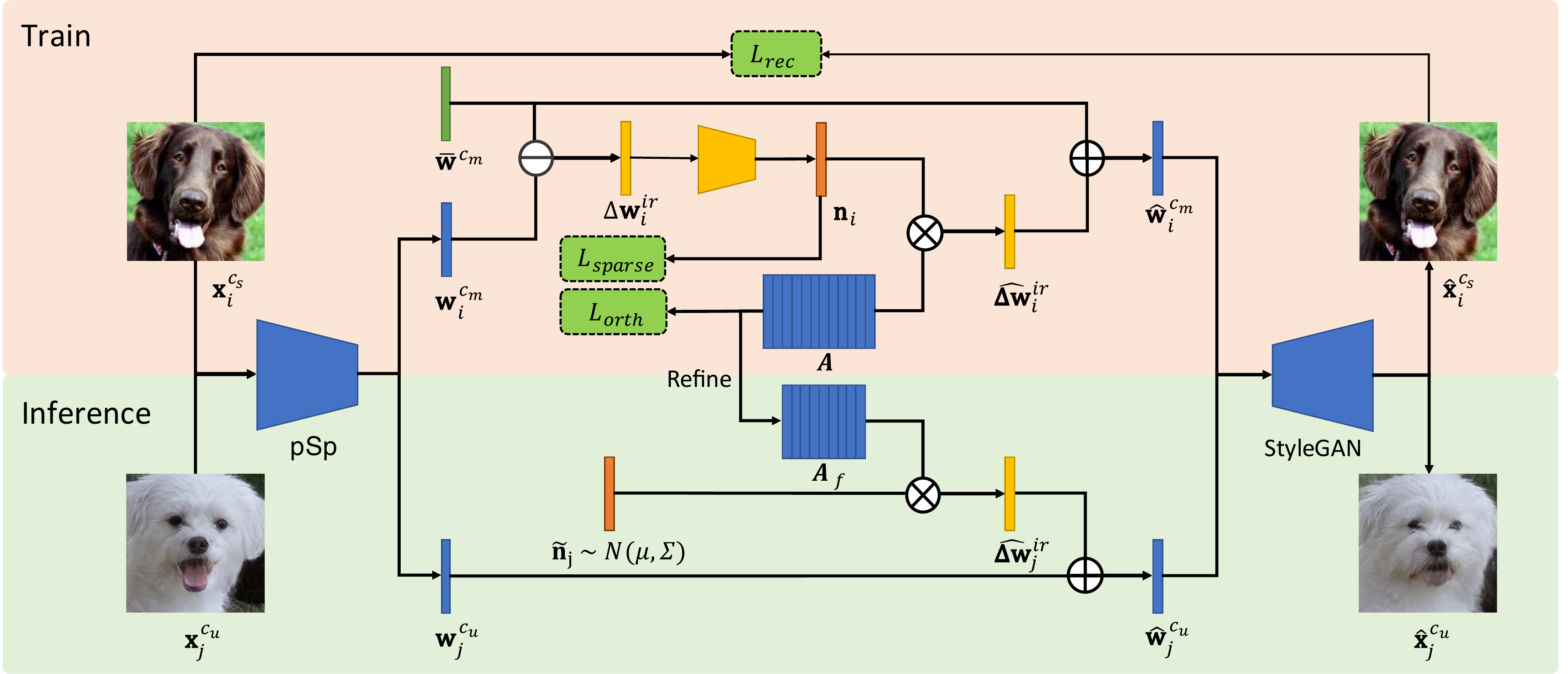}
   
    \caption{The overview of AGE. AGE learns a dictionary $\bm{A}$ consisting of category-irrelevant attribute editing directions in the training stage and generates diverse images via category-irrelevant manipulation according to the refined dictionary $\bm{A}_f$. }
    \vspace{-1em}
    \label{architechture}
\end{figure*}

\noindent\textbf{Image manipulation.} Recent studies have shown that GANs can represent multiple interpretable attributes in the latent space~\cite{stylegan, sehier}. 
For image editing, supervised learning methods~\cite{ganalyze, sehier, interfacegan} annotate predefined attributes according to pre-trained classifiers, and then learn the potential directions of the attributes in the latent space. However, they heavily rely on the attribute predictors and human annotations.
Some concurrent work studies unsupervised semantic discovery in GANs. 
The meaningful dimensions can be identified by using segmentation-based networks~\cite{RW11-GANDissect}, linear subspace models~\cite{eigengan}, Principal Components Analysis in the activation space~\cite{RW11-GANSpace}, or carefully designed disentanglement constraints~\cite{sefa,RW11-Hessian,jacobian}.
Different from traditional image editing, AGE focuses on attribute factorization on more challenging multi-category image generation, which cannot be performed by traditional image editing methods.



\section{Method}

A training set $\mathcal{D}_{train} = \{x_i^{c_m}\}^{N_m \times M}$ consists of $M$ seen categories and a testing set $\mathcal{D}_{test}= \{x_i^{c_k}\}^{N_k \times K}$ consists of $K$ unseen categories, where the number of images $N_k$ in each category is generally small, {\it i.e.}, 10 or 15. Few-shot image generation aims to train a multi-category generative network with $\mathcal{D}_{train}$ but generates diverse images of the $K$ unseen categories by the few images in $\mathcal{D}_{test}$.

In this section, we introduce Attribute Group Editing (AGE), a method to generate images of unseen categories without re-training a GAN model. AGE makes use of a large number of images of known categories to identify the semantically meaningful directions of category-relevant and category-irrelevant attributes without explicit supervision. 

\label{sec:formatting}
\subsection{Preliminaries}
\noindent\textbf{GAN Inversion.} The generator $G(\cdot)$ in GANs learns a mapping from the $d$-dimensional latent space $\mathcal{Z}\in \mathbb{R}^d$ to higher dimensional images $\mathcal{X}\in \mathbb{R}^{H\times W\times C}$. On the contrary, given an image $x_i\in \mathcal{X}$, it can also be embedded to the latent space with GAN inversion~\cite{psp, bdinvert, ganinversion, wplus, wplus1}. The process of GAN inversion $I(\cdot)$ and generation can be formulated as follow:
\begin{equation}
  \mathbf{z}_i=I(x_i), \ \  \hat{x}_i=G(\mathbf{z}_i).
  \label{eq:gen}
\end{equation}

\noindent\textbf{Semantic Manipulation.} The latent space of GANs has recently been shown to encode rich semantic knowledge~\cite{ganalyze, sehier, steer}. Different directions in the latent space control different attributes. Many works~\cite{sefa, eigengan, interfacegan, jacobian} proposed to manipulate the latent vector $\mathbf{z}_i$ in a certain direction $\Delta\mathbf{z}_i\in \mathbb{R}^d$ to edit the corresponding attribute:
\begin{equation}
  \texttt{edit}(G(\mathbf{z}_i)) = G(\mathbf{z'}_i) = G(\mathbf{z}_i + \alpha\Delta\mathbf{z}_i),
  \label{eq:important}
\end{equation}
where $\texttt{edit}(\cdot)$ denotes the editing operation on images. $\alpha$ stands for the manipulation intensity.

Given a well-trained GAN for multi-category image generation, editing can be divided into category-relevant editing and category-irrelevant editing. 
For a sampled latent vector $\mathbf{z}_i^{c_m}$ of the category $c_m$,  category-relevant editing $\texttt{edit}_r(\cdot)$ is:
\begin{equation}
  \texttt{edit}_r(G(\mathbf{z}_i^{c_m})) = G(\mathbf{z}_i^c + \alpha\Delta\mathbf{z}^r) = \hat{x}_i^{c_k},
  \label{eq:relevant}
\end{equation}
where $\Delta\mathbf{z}^r$ denotes the directions of category-relevant manipulation and $\hat{x}_i^{c_k}$ is an image of a new category $c_k$. 

On the other hand, for category-irrelevant editing $\texttt{edit}_{ir}(\cdot)$, we have:
\begin{equation}
  \texttt{edit}_{ir}(G(\mathbf{z}_i^{c_m})) = G(\mathbf{z}_i^{c_m} + \alpha\Delta\mathbf{z}^{ir}) = \hat{x}_i^{c_m},
  \label{eq:ir}
\end{equation}
where $\Delta\mathbf{z}^{ir}$ denotes the directions of category-irrelevant editing and $\hat{x}_i^{c_m}$ is an image of the same category $c_m$. 

\begin{figure}[ht]
	\centering
	\includegraphics[width=0.95\linewidth]{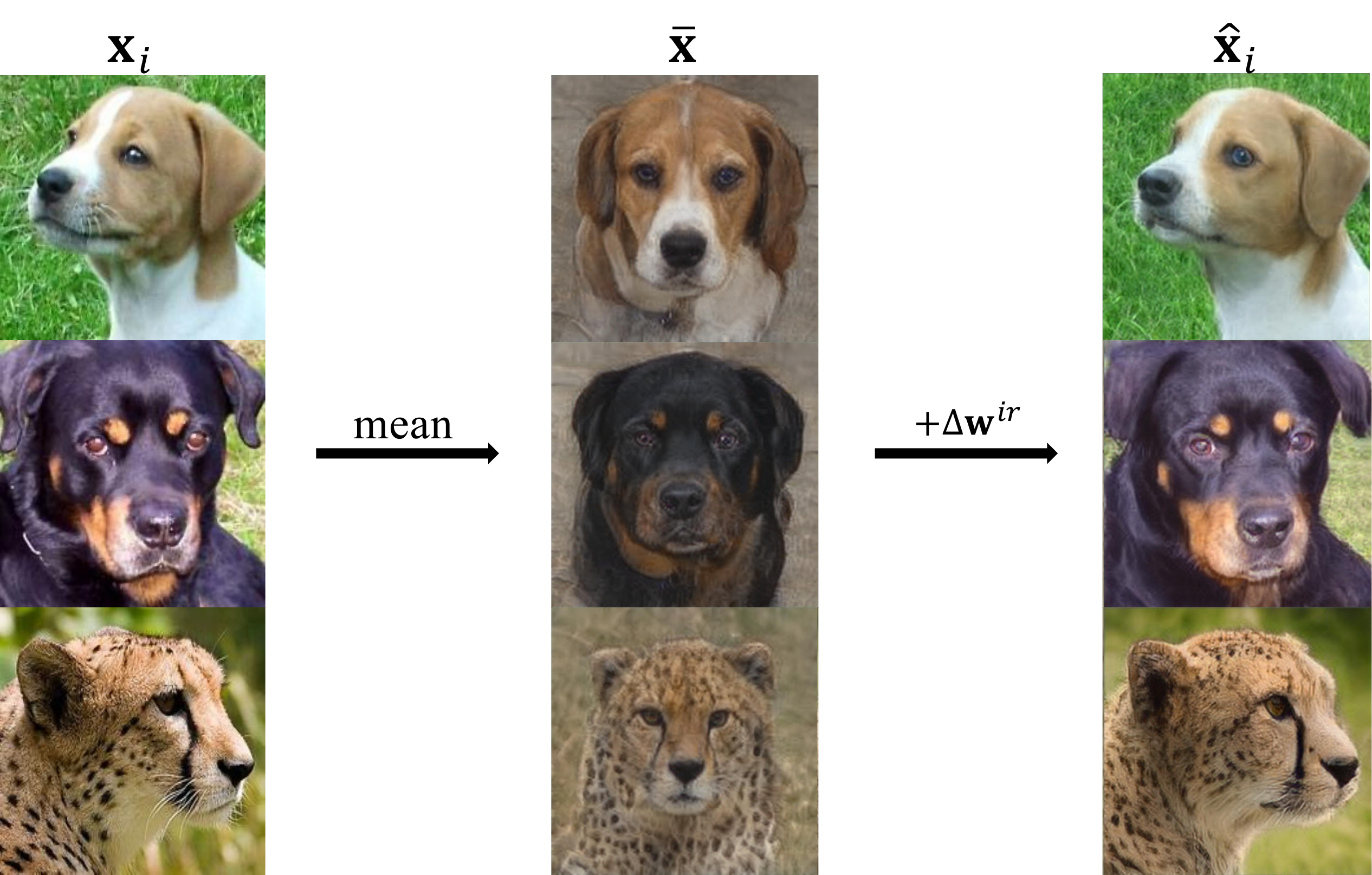}
	
	\caption{Any image can be regarded as category-irrelevantly edited from the corresponding class embedding. This linear transformation can reconstruct most attributes of the input images.}
	\vspace{-1em}
	\label{recons}
\end{figure}

\subsection{Attribute Group Editing}
The overall framework of AGE is shown in Figure~\ref{architechture}, which consists of two main parts: Image Embedding and Attribute Factorization.

\subsubsection{Image Embedding}
To achieve image editing, we should first embed the image samples into the latent space.
In practice, we employ pSp~\cite{psp} to encode an image $x_i$ to $\mathcal{W^+}$~\cite{wplus1} space of StyleGAN2~\cite{stylegan, stylegan2}. 
\begin{equation}
  \mathbf{w}_i=\texttt{pSp}(x_i),
  \label{eq:psp}
\end{equation}
where $\mathbf{w}_i\in\mathbb{R}^{18\times512}$ is the corresponding latent vector of $x_i$ in the $\mathcal{W^+}$ space. 

\subsubsection{Attribute Factorization}
Given latent representations in $\mathcal{W}^+$ space, we aim to distinguish a set of category-relevant attribute directions and category-irrelevant attribute directions according to the given dataset. 

\noindent\textbf{Category-relevant Attributes}.
The combination of category-relevant attributes identifies which category an image belongs to.
It encodes the common attributes among all samples from one specific category.
Therefore, we use the mean vector $\overline{\mathbf{w}}^{c_m}\in\mathcal{W^+}$ of all samples in a category $c_m$ to represent the class embedding, \ie, the combined category-relevant attributes of $c_m$:
\begin{equation}
    \overline{\mathbf{w}}^{c_m} = \frac{1}{N_m} \sum_{i=1}^{N_m} \mathbf{w}_i^{c_m},
\end{equation}
where $N_m$ is the number of samples from the category $c_m$.

The dictionary of category-relevant attributes of $M$ seen categories is defined as $\bm{B}=[\overline{\mathbf{w}}^{c_1},\overline{\mathbf{w}}^{c_2},...,\overline{\mathbf{w}}^{c_M}]$.


\noindent\textbf{Category-irrelevant Attributes.} 
As shown in Figure~\ref{recons}, given $\overline{\mathbf{w}}^{c_m}$ contains all category-relevant attributes, any image $x_i^{c_m}$ of the category $c_m$ can be obtained by manipulating with category-irrelevant editing $\Delta\mathbf{w}^{ir}$ as Eq.~\ref{eq:dw}:
\begin{equation}
  \mathbf{w}_i^{c_m}=\overline{\mathbf{w}}^{c_m}+\Delta\mathbf{w}_i^{ir}.
  \label{eq:dw}
\end{equation}

The category-irrelevant directions are common across all known and unknown categories. 
To learn the category-irrelevant directions, we model this manipulation process as Sparse Dictionary Learning (SDL)~\cite{sdl, ksvd}. 

Given a sample of manipulation direction $\Delta\mathbf{w}_i^{ir}$, we optimize a global dictionary $\bm{A}\in\mathbb{R}^{18\times512\times l}$ that contains all directions of category-irrelevant attributes and a sparse representation $\mathbf{n}_i \in \mathbb{R}^{18\times l}$ with
\begin{equation}
    \min_{\mathbf{n}} \|\mathbf{n}_i\|_0 \ \  \ \ \text{s.t.} \ \  \Delta\mathbf{w}_i^{ir}=\bm{A}\mathbf{n}_i, 
  \label{eq:sdl}
\end{equation}
where $\|.\|_0$ is the $L_0$ constraint.
This sparse constraint encourages each element in $\bm{A}$ to be semantically meaningful. 

In practice, it is optimized via an Encoder-Decoder architecture.
The sparse representation $\mathbf{n}_i$ is obtained from $\Delta\mathbf{w}_i^{ir}$ with a Multi-layer Perceptron (MLP):
\begin{equation}
  \mathbf{n}_i = \texttt{MLP}(\Delta\mathbf{w}_i^{ir}).
  \label{eq:}
\end{equation}

Since the $L_0$ loss is not derivable, we approximate $L_0$ constraint with $L_1$ with the sigmoid activation:
\begin{equation}
  L_{\text{sparse}} = \|\sigma(\theta_0\mathbf{n}_i-\theta_1)\|_1,
  \label{eq:reconstruct}
\end{equation}
where $\sigma(\cdot)$ denotes the sigmoid function. $\theta_0$ and $\theta_1$ are hyper-parameters to control the sparsity.

The generator is to generate an image close to the input $x_i$, which is optimized with the $L_2$ reconstruction loss:
\begin{equation}
 L_{\text{rec}} = \|G(\mathbf{\overline{w}}^{c_m}+A\mathbf{n}_i)- x_i^{c_m}\|_2.
  \label{eq:important}
\end{equation}

Moreover, to further guarantee that $\bm{A}\mathbf{n}_i$ only edits category-irrelevant attributes, the embedding of an edited images $\mathbf{\hat{w}}^{c_m}_i$ should have the same category-relevant attributes response as the class embedding $\mathbf{\overline{w}}^{c_m}$:
\begin{equation}
\begin{aligned}
  \bm{B}^T\mathbf{\hat{w}}^{c_m}_i&=\bm{B}^T\mathbf{\overline{w}}^{c_m}, \\
  \bm{B}^T\mathbf{\overline{w}}^{c_m}+\bm{B}^T\bm{A}\mathbf{n}_i&= \bm{B}^T\mathbf{\overline{w}}^{c_m} ,\\
  \bm{B}^T\bm{A}\mathbf{n}_i&=\bm{0}.
  \label{eq:ab}
\end{aligned}
\end{equation}

To ensure the satisfaction of Eq.~\ref{eq:ab}, we formulate an orthogonal constraint between $\bm{A}$ and $\bm{B}$ with:
\begin{equation}
  L_{\text{orth}} = \|\bm{B}^T\bm{A}\|_F^2,
  \label{eq:oth}
\end{equation} 
where $\|.\|_F^2$ denotes the Frobenius Norm.

The overall loss function is
\begin{equation}
  L = L_{\text{rec}} + \lambda_1 L_{\text{orth}} + \lambda_2 L_{\text{sparse}}.
  \label{eq:oth}
\end{equation} 

\begin{figure*}[ht]
    \centering
    \includegraphics[width=0.85\linewidth]{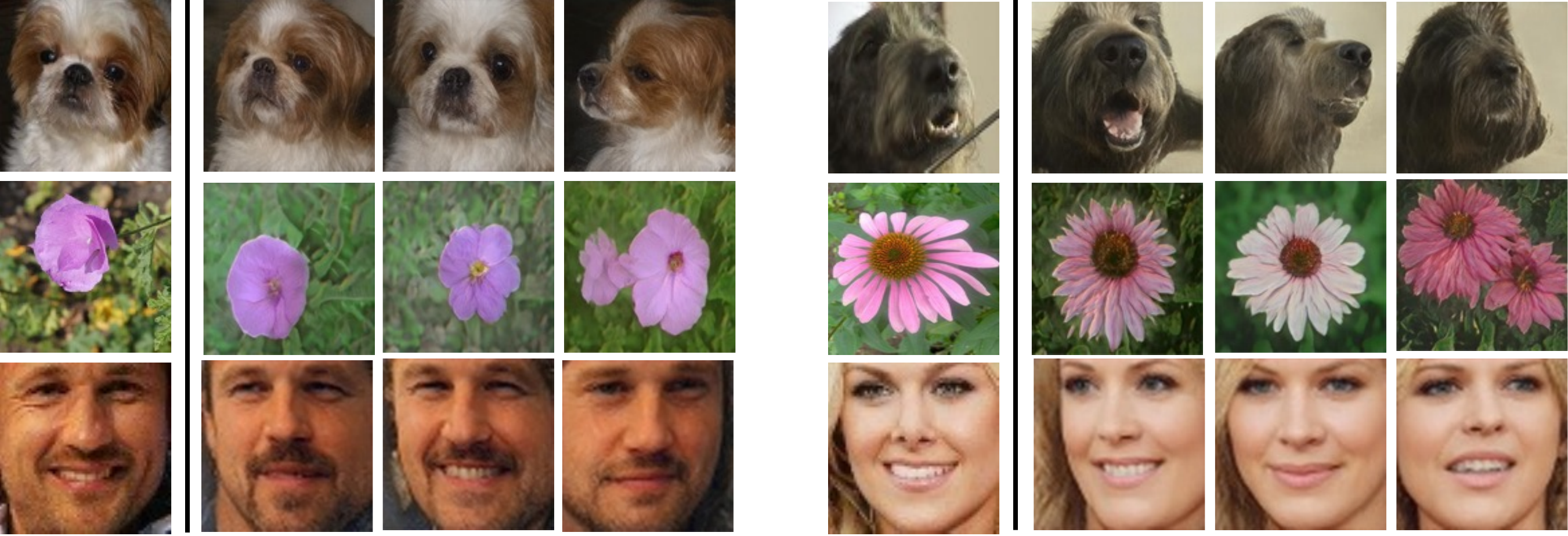}
   \vspace{-0.5em}
    \caption{One-shot image generation from AGE on Animal Faces, Flowers and VGGFaces.}
    \label{fig:good}
\end{figure*}

\begin{table*}[t]
	\centering
	\setlength{\tabcolsep}{4.5mm}
	\caption{FID($\downarrow$) and LPIPS($\uparrow$) of images generated by different methods for unseen categories. VGGFaces is marked with * because most methods report different number of unseen categories on this dataset (\eg 552 in LoFGAN, 96 in DeltaGAN and 497 in L2GAN).}
    \begin{tabular}{lccccccc}
    \hline
    \multirow{2}{*}{Method} & \multirow{2}{*}{Settings} & \multicolumn{2}{c}{Flowers}      & \multicolumn{2}{c}{Animal Faces} & \multicolumn{2}{c}{VGG Faces$^*$}    \\
                            &                           & FID(↓)         & LPIPS(↑)        & FID(↓)         & LPIPS(↑)        & FID(↓)         & LPIPS(↑)        \\ \hline
    FIGR~\cite{figr}                    & 3-shot                    & 190.12         & 0.0634          & 211.54         & 0.0756          & 139.83         & 0.0834          \\
    GMN~\cite{gmn}                     & 3-shot                    & 200.11         & 0.0743          & 220.45         & 0.0868          & 136.21         & 0.0902          \\
    DAWSON~\cite{dawson}                  & 3-shot                    & 188.96         & 0.0583          & 208.68         & 0.0642          & 137.82         & 0.0769          \\
    DAGAN~\cite{dagan}                   & 1-shot                    & 179.59         & 0.0496          & 185.54         & 0.0687          & 134.28         & 0.0608          \\
    MatchingGAN~\cite{matchinggan}             & 3-shot                    & 143.35         & 0.1627          & 148.52         & 0.1514          & 118.62         & 0.1695          \\
    F2GAN~\cite{f2gan}                   & 3-shot                    & 120.48         & 0.2172          & 117.74         & 0.1831          & 109.16         & 0.2125          \\
    LoFGAN~\cite{lofgan}                  & 3-shot                    & 79.33          & 0.3862          & 112.81         & 0.4964          & \textbf{20.31} & 0.2869          \\
    DeltaGAN~\cite{deltagan}                & 1-shot                    & 109.78         & 0.3912          & 89.81          & 0.4418          & 80.12 & 0.3146 \\ \hline
    AGE                     & 1-shot                    & \textbf{45.96} & \textbf{0.4305} & \textbf{28.04} & \textbf{0.5575} & 34.86          & \textbf{0.3294} \\ \hline
    \end{tabular}
	\vspace{-0.5em}
\label{tab:fid}
\end{table*}

In the inference phase, in order to find out the most common category-irrelevant editing directions,
we first back-project $\Delta\mathbf{w}^{ir}$ onto the representation $\mathbf{\hat{n}}$:
\begin{equation}
  \mathbf{\hat{n}}_i=\bm{A}^{-1}\Delta\mathbf{w}^{ir}_i,
  \label{eq:backpro}
\end{equation} 
where $\bm{A}^{-1}$ is the pseudo-inverse matrix of $\bm{A}$. Afterward, we count the mean of the absolute value of $|\mathbf{\hat{n}}_i|$ across all $M$ seen categories:
\begin{equation}
  \mathbf{\overline{|\hat{n}|}} = \frac{1}{M} \sum_{m=1}^{M} \frac{1}{N_m} \sum_{i=1}^{N_m}  |\mathbf{\hat{n}}_i^{c_m}|,
  \label{eq:mean_n}
\end{equation} 
where $\mathbf{\overline{|\hat{n}|}}$ can be interpreted as the commonality of the directions across the whole dataset. For each layer of the $\mathcal{W}^+$ space, we select $t$ directions from $\bm{A}$ that correspond to top-$t$ values in $\bm{\overline{|\hat{n}|}}$. The final dictionary for category-irrelevant editing is $\bm{A}_f\in\mathbb{R}^{18\times512\times t}$.


To automatically generate diverse images, we assume that the sparse representation $\mathbf{n}$ obeys a Gaussian distribution $\mathcal{N}(\mu, \Sigma)$, which is obtained by counting the $\hat{\mathbf{n}}_i$ of all seen categories in the training set. We sample an arbitrary $\tilde{\mathbf{n}}_j$ from $\mathcal{N}(\mu, \Sigma)$ and apply editing to unseen category images. The manipulation intensity $\alpha$ is introduced to control the diversity of generated images. Given a single image $x_i^{c_n}$, a set of images can be generated with
\begin{equation}
    x_j^{c_k} = G(\mathbf{w}_i^{c_k} + \alpha \bm{A}_f \tilde{\mathbf{n}}_j),
\end{equation}
where $\mathbf{w}_i^{c_k} = \texttt{pSp}(x_i^{c_k})$.

\section{Experiment}
\label{sec:exp}
\subsection{Implementation Details}
In the training stage, we first train a StyleGAN2~\cite{stylegan2} with seen categories. 
Given a trained GAN, the sparse representation encoder is a 5-layer multi-layer perceptron with Leaky-ReLU activation function. The length  $l$ of dictionary $\bm{A}$ is set to $100$. 
For more stable and interpretable editing, we group the 18-layers $\mathcal{W}^+$ space of StyleGAN2 into bottom layers, middle layers, and top layers, corresponding to 0-2, 3-6, and 7-17 layers respectively. Layers in each group share the same sparse representation $\mathbf{n}$.


\begin{figure*}[t]
    \vspace{-1em}
	\centering	
	\includegraphics[width=1\linewidth]{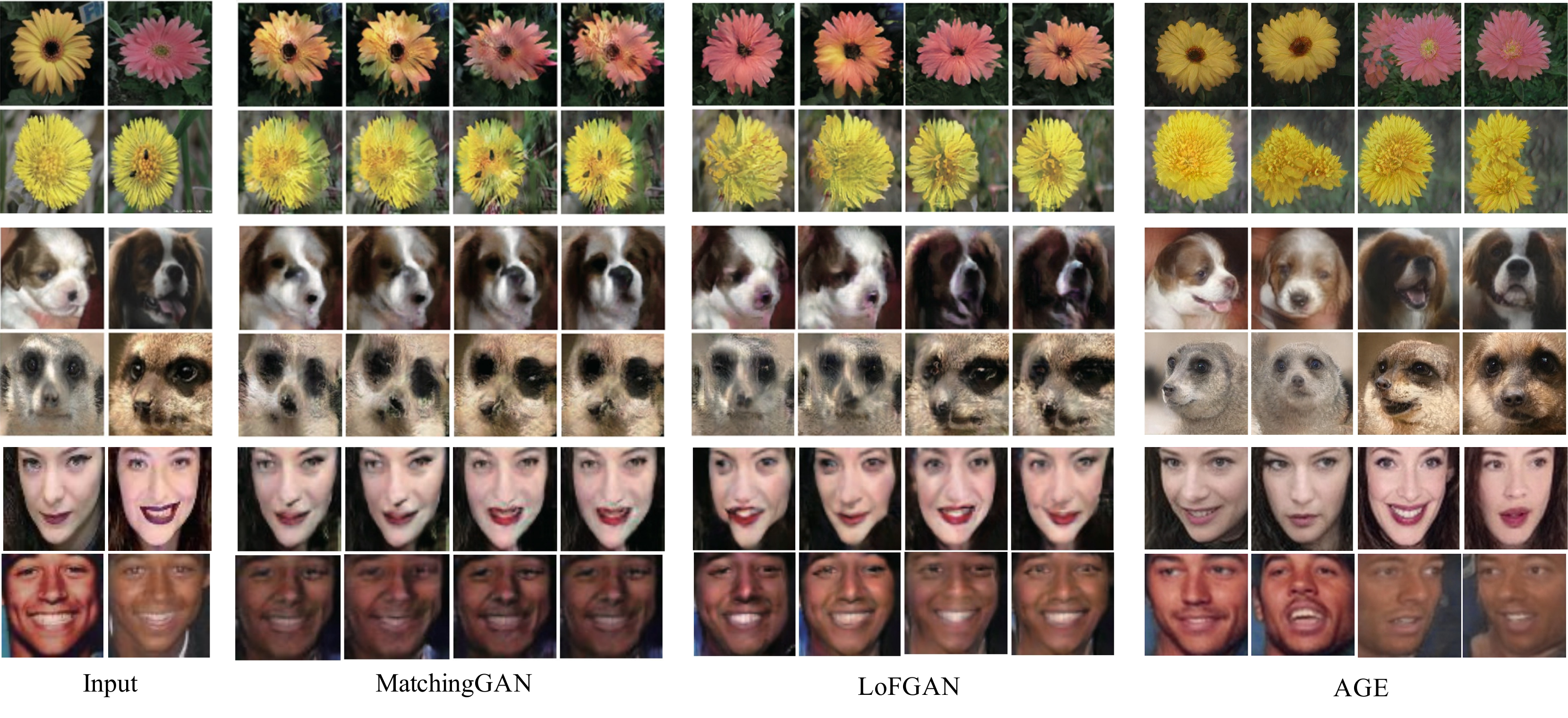}
	\vspace{-2em}
	\caption{Comparison between images generated by MatchingGAN, LoFGAN, and AGE on Flowers, Animal Faces, and VGGFaces.}
	\vspace{-1em}
	\label{fig:compare}
\end{figure*}

\subsection{Datasets}
We evaluate our method on Animal Faces~\cite{animalfaces}, Flowers~\cite{flowers}, and VGGFaces~\cite{vggfaces} following the settings in \cite{deltagan}.

\noindent\textbf{Animal Faces}. We select 119 categories as seen categories for training and 30 as unseen categories for testing. 

\noindent\textbf{Flowers}. We split it into 85 seen categories for training and 17 unseen categories for testing. 

\noindent\textbf{VGGFaces}. For VGGFaces~\cite{vggfaces}, we randomly select 1802 categories for training and 572 for evaluation.

\subsection{Ablation Study on Downstream Task}
We test data augmentation for image classification on Animal Faces~\cite{animalfaces}. We randomly select 15, 35, 100 images for each category as train, val, and test, respectively. Following \cite{lofgan}, 
a ResNet-18 backbone is first initialized from the seen categories, then the model is fine-tuned on the unseen categories. 75 images are generated for each unseen category as data augmentation. 

\begin{figure}[t]
	\centering	
	\includegraphics[width=1\linewidth]{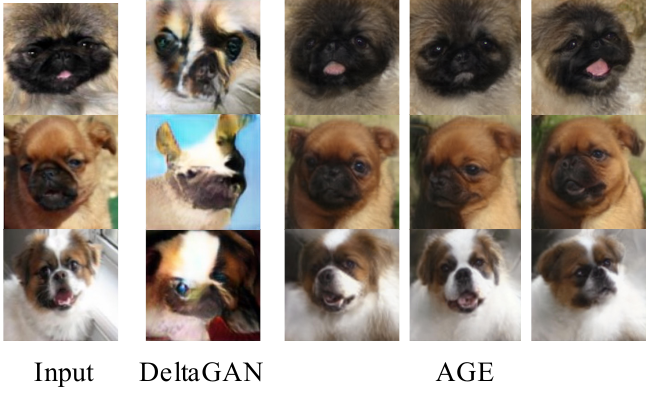}
	\vspace{-2em}
	\caption{Many failure cases in DeltaGAN can be stably generated by AGE.}
	\vspace{-1.4em}
	\label{fig:delta}
\end{figure}

``Sample Train" is an ablation that randomly samples $\Delta\mathbf{w}$ of seen categories from the train set and directly used to edit the unseen categories.
As shown in Table~\ref{tab:ab}, the directly sampled $\Delta\mathbf{w}$ is unstable for image editing, resulting in crashed images and much higher FID, which proves the necessity of the attribute factorization with SDL. 

\begin{figure*}[t]
	\centering
	\includegraphics[width=0.9\linewidth]{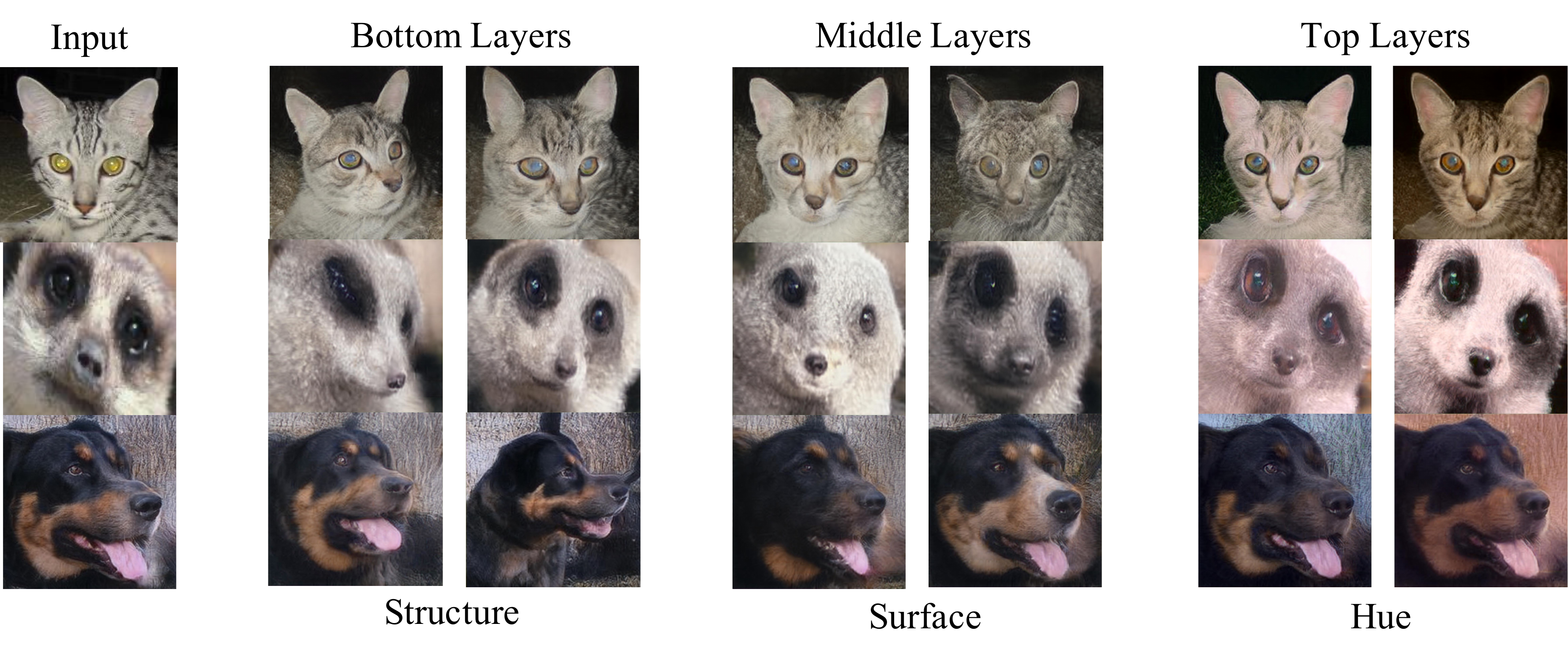}
	\vspace{-1.5em}
	\caption{{ Hierarchical interpretable manipulations discovered in $\bm{A}_f$.}}
	\vspace{-1em}
	\label{fig:group}
\end{figure*}

The diversity and quality of generated images are largely controlled by the manipulation intensity $\alpha$.
With the growth of $\alpha$, AGE generates images of higher diversity but lower quality. According to the accuracy, $\alpha=1$ achieves the best performance on the classification. This result indicates that generating images that follow the same distribution with the original training set is the best choice for data augmentation. Although lower $\alpha$ can achieve more satisfactory FID and LPIPS, we still decide the best $\alpha$ according to the performance on downstream tasks.

\begin{table}[t]
\centering
\setlength{\tabcolsep}{4.3mm}
\caption{Ablations of different manipulation intensity $\alpha$.}
\begin{tabular}{@{}cccc@{}}
\toprule
$\alpha$ & Accuracy      & FID ($\downarrow$)        & LPIPS ($\uparrow$)                      \\ \midrule
Baseline   & 67.3          & --                                 & --                        \\ \midrule
Sample Train   &    70.8       & 54.66   &   0.6103   \\ \midrule
0.3        & 69.8          & \textbf{31.79} & 0.5429\\
0.5        & 70.6          & 31.81                              & 0.5482                              \\
0.7        & 70.9          & 33.43                              & 0.5532                              \\
1.0        & \textbf{71.4} & 38.18   & 0.5609     \\
1.5        & 69.9          & 49.70                              & 0.5719                              \\
2.0        & 66.1          & 63.99       & \textbf{0.5809} \\ \bottomrule
\end{tabular}
\label{tab:ab}
\vspace{-1em}
\end{table}

\subsection{Quantitative Comparison with State-of-the-art}

We evaluate the quality of the generated images based on commonly used FID and LPIPS.
Following the former works~\cite{lofgan, matchinggan, deltagan}, we generate 128 images based on sampled real images of each unseen category and calculate FID and LPIPS based on the generated images. 
Following one-shot settings in ~\cite{dagan, deltagan}, one real image is used each time to generate adequate images for unseen categories. 

The results of different methods are reported in Table~\ref{tab:fid}, our method achieves significant improvements on both FID and LPIPS. Since we do not need to re-train a GAN, AGE is much more stable, achieving impressive FID gain. Compared with fusion-based and transformation-based methods, our generated images are also more diverse. Besides, we can achieve one-shot image generation.

\begin{figure}[t]
	\centering
	\includegraphics[width=0.9\linewidth]{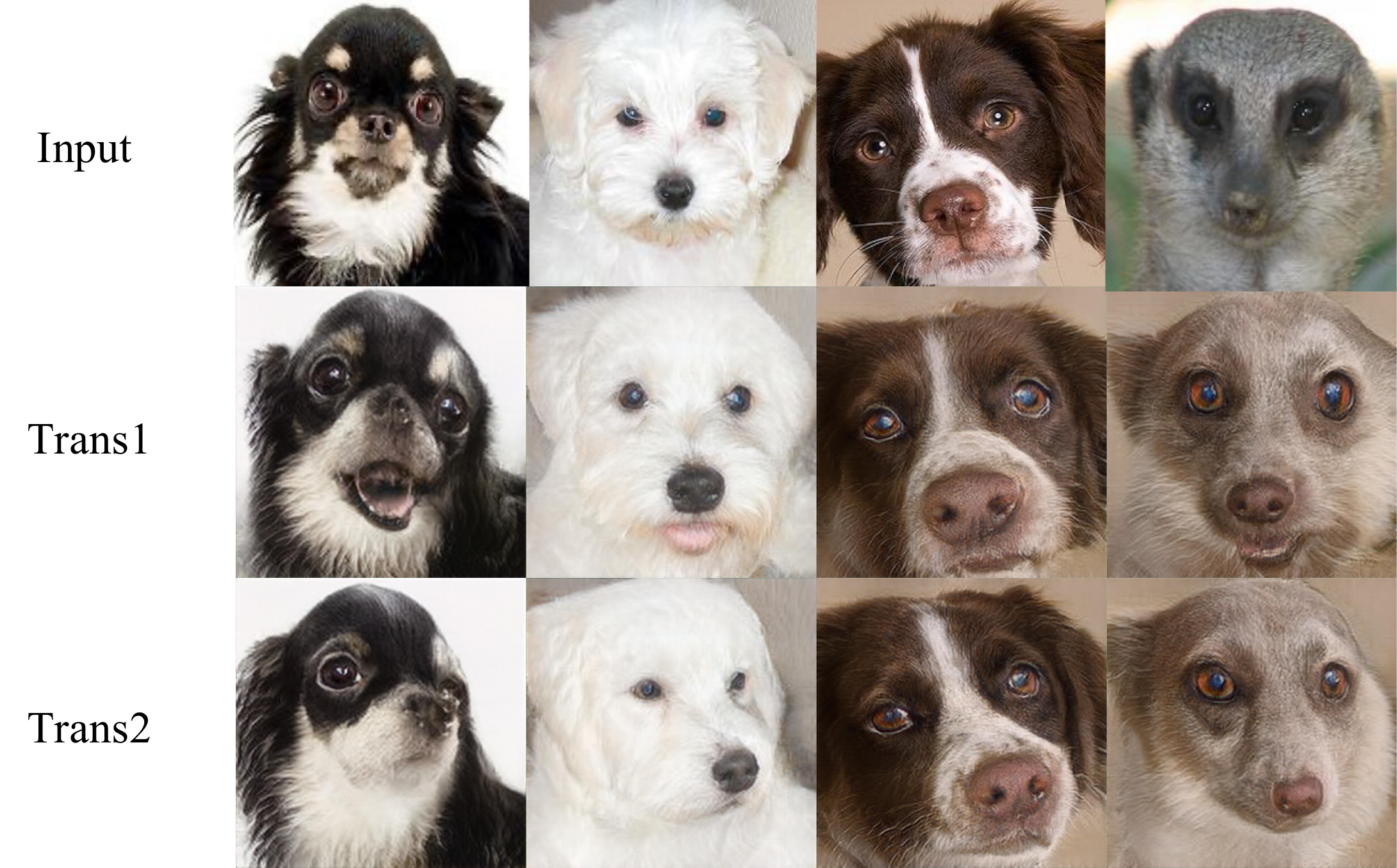}
	\vspace{-0.5em}
	\caption{{Manipulate images from different categories with the same sparse representation $\mathbf{n}$}.}
	\label{fig:trans}
	\vspace{-1.5em}
\end{figure}

\subsection{Qualitative Evaluation}
One-shot image generation from AGE on Animal Faces, Flowers and VGGFaces is shown in Figure~\ref{fig:good}.
We qualitatively compare our method with the fusion-based methods MatchingGAN~\cite{matchinggan} and LoFGAN~\cite{lofgan}, and transformation-based method DeltaGAN~\cite{deltagan}. All images are reported in~\cite{lofgan} and~\cite{deltagan}. 
As shown in Figure~\ref{fig:compare}, different from LoFGAN~\cite{lofgan} that can only fuse features from conditional images, AGE can produce images that have new attributes. For example, we can generate images with two flowers and generate dogs of diverse positions and expressions. 

Compared with the transformation-based method DeltaGAN that learns intra-category transformations from different image pairs, AGE obtains the latent category-irrelevant semantics based on more robust class embeddings. Many failure cases in DeltaGAN can be stably generated with AGE as shown in Figure~\ref{fig:delta}.
Moreover, since AGE does not retrain a GAN, the generated images from AGE are of much higher quality compared with existing fusion-based and transformation-based methods.

\subsection{Semantic Attribute Factorization}
Apart from few-shot image generation, an additional advantage for AGE is controllable image editing about the category-irrelevant attributes. 
In this section, we will experimentally demonstrate the transferability and interpretability of the learned dictionary $\bm{A}_f$.

\noindent\textbf{Transferability}. 
Since $\bm{A}_f$ is category-irrelevant, it is transferable across all 
categories. We edit the images from 4 categories with the same editing direction, the output images are shown in Figure~\ref{fig:trans}. 
\texttt{Trans1} is to open the mouth and \texttt{Trans2} is to turn the head to right. It demonstrates that the dictionary $\bm{A}_f$ is global and the same $\mathbf{n}$ controls similar attributes for images of different categories. 

\begin{figure}[t]
	\centering
	\includegraphics[width=\linewidth]{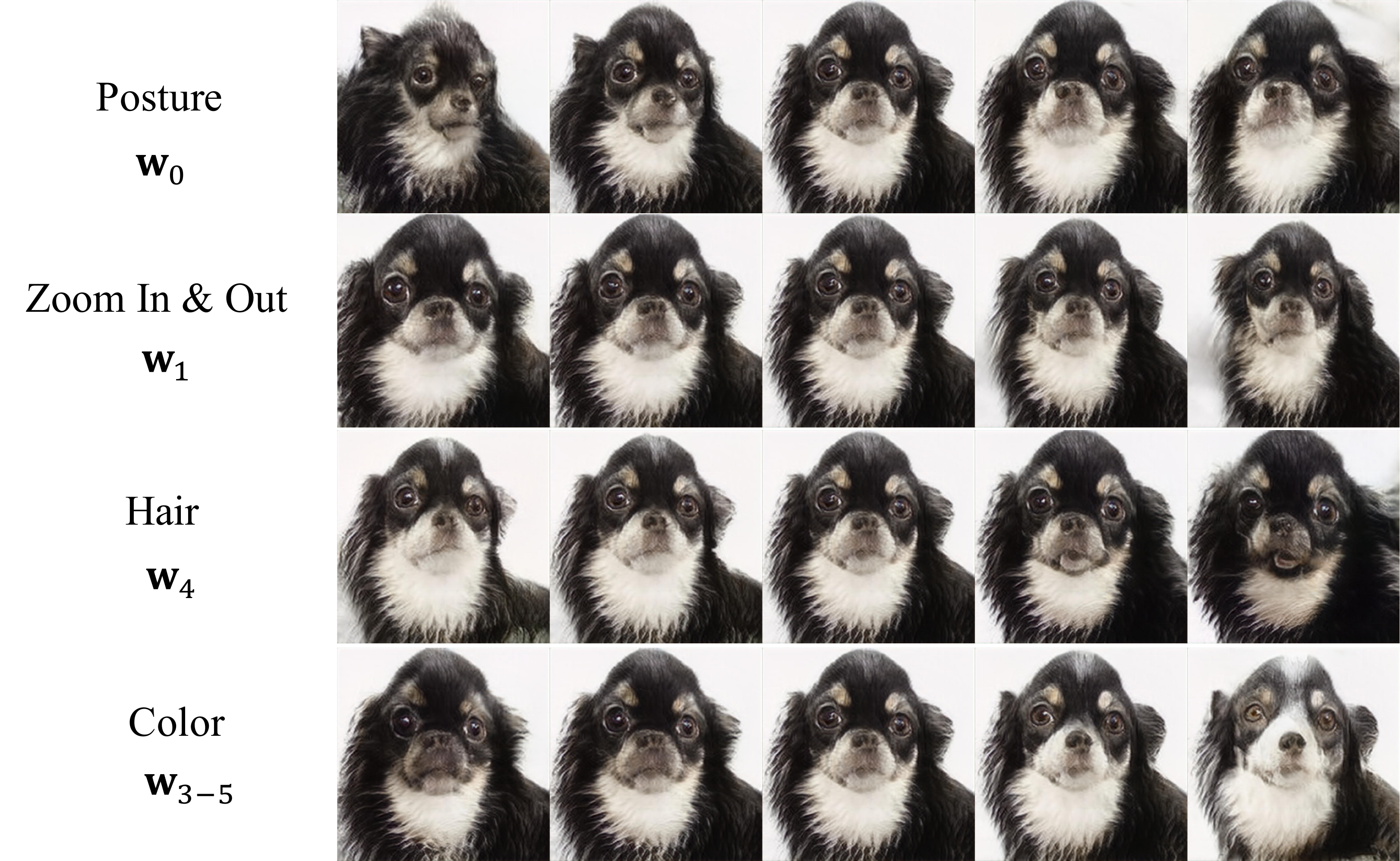}
	\caption{Manipulate images along disentangled attribute editing directions in different layers in $\bm{U}$.}
	\vspace{-0.8em}
	\label{fig:svd}
\end{figure}

\noindent\textbf{Interpretability}. 
Although AGE is performed in a completely self-supervised manner, the learned directions in $\bm{A}_f$ are still interpretable due to the sparse constraint and the meaningful latent space of StyleGAN2.

First, different groups of layers control different category-irrelevant attributes. In particular, we interpret a target model at the levels of bottom layers, middle layers, and top layers, respectively. Figure~\ref{fig:group} shows the versatile semantic directions found in Animal Faces. 
It demonstrates that most directions in $\bm{A}_f$ is category-irrelevant.
Concretely, the bottom layers mainly control the structure of objects, such as the position, zoom in/out, and the shape of the face. The middle layers mainly control the surface features like color and the expressions. The top layers decide the background and the overall hue of the image. We can achieve controllable category-irrelevant editing by sampling in corresponding groups.

To find the disentangled attribute editing directions in different layers/groups, we further conduct Singular Value Decomposition (SVD) on dictionary $\bm{A}_f$:
\begin{equation}
    \bm{A}_f = \bm{U \Sigma V^*}.
\end{equation}
Matrix $\bm{U}$ contains the commonly shared directions of each layer of dictionary $\bm{A}_f$. Figure~\ref{fig:svd} is an illustration of image editing by moving along the most salient directions of each layer. Although the disentanglement is rough, moving along a single direction in $\bm{U}$ can continuously edit one specific category-irrelevant attribute.

\subsection{Ablation Study of Loss Components}
\begin{figure}[t]
\centering
	\setlength{\abovecaptionskip}{0cm}
	\includegraphics[width=0.9\linewidth]{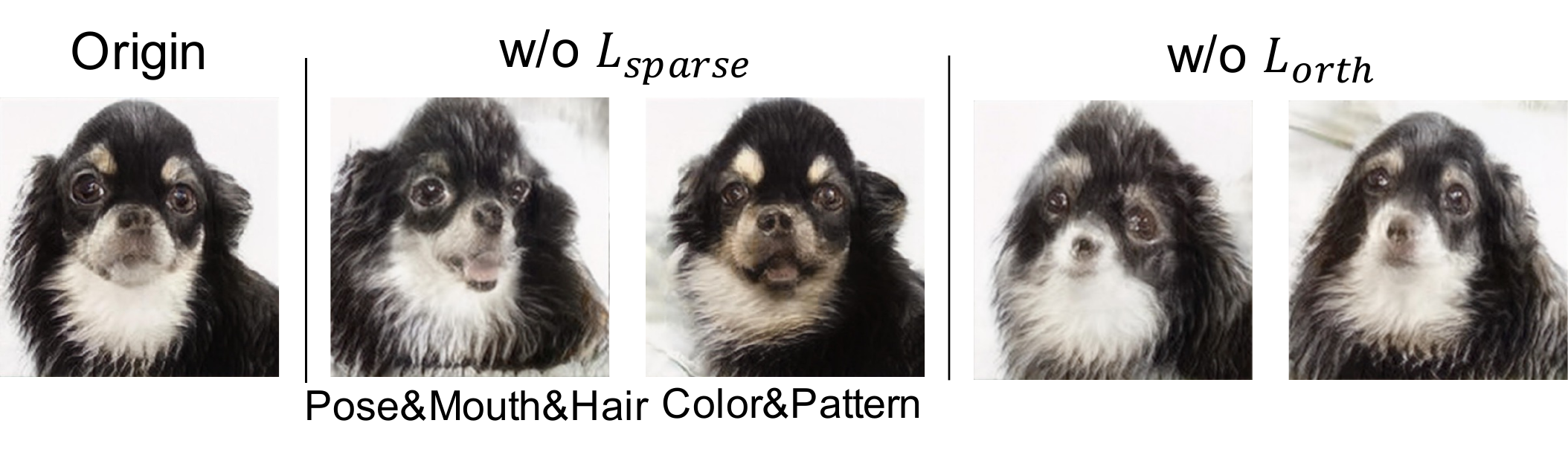}
	\caption{Image editing without $ L_{\text{sparse}}$ or $ L_{\text{otrh}}$}
	\vspace{-1.5em}
	\label{fig:absvd}
\end{figure}

We conduct an ablation study on $L_{\text{orth}}$ and $ L_{\text{sparse}}$ on Animal Faces~\cite{animalfaces}. 
$ L_{\text{orth}}$ is to encourage category-irrelevance of the learned directions. The edited images are more likely to encounter category change without $ L_{\text{orth}}$ in Figure~\ref{fig:absvd}. As a result, the accuracy of few-shot image classification is obviously lower than AGE without $L_{\text{orth}}$  as shown in Table~\ref{tab:ab}.
$ L_{\text{sparse}}$ is for sparsity of the representation $\mathbf{n}$, which is related to the interpretability of the learned directions.
Compared with Figure~\ref{fig:svd}, it shows that different semantics are more entangled in the learned directions after SVD without $ L_{\text{sparse}}$. Editing image along one direction will cause the change of several attributes.

\begin{table}[t]
    \caption{Ablation of $L_{\text{orth}}$ on few-shot image classification.}
    \centering
    \setlength{\tabcolsep}{4mm}
    \begin{tabular}{lccc}
    \hline
    Setting  & baseline & w/o $L_{\text{orth}}$ & AGE  \\ \hline
    Accuracy & 67.3     & 69.8                  & 71.4 \\ \hline
    \end{tabular}
    \vspace{-0.8em}
    \label{tab:ab}
\end{table}

\subsection{Failure Cases and Limitations}
\begin{figure}[t]
	\centering
	\includegraphics[width=\linewidth]{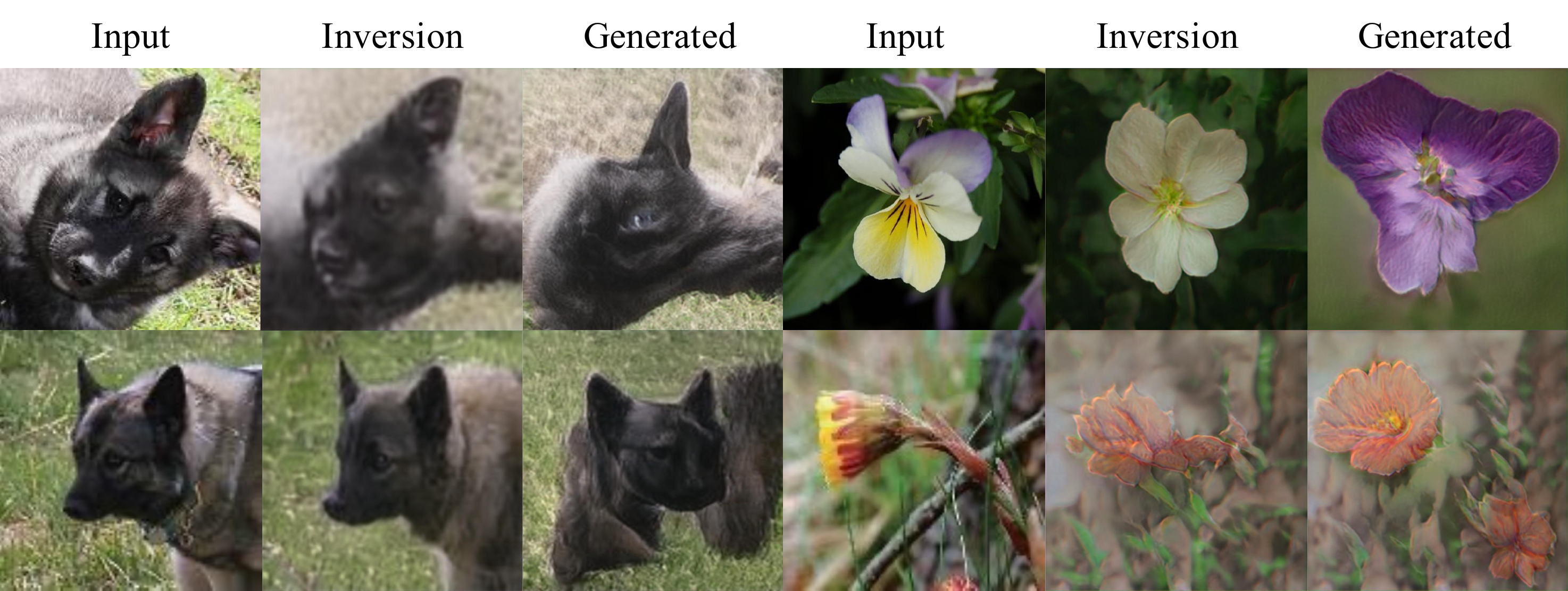}
	\vspace{-1em}
	\caption{{\bf Failure Cases}. The failure cases majorly comes from bad GAN inversion, irregular positions of input objects, and wrong category-relevant editing. }
	\vspace{-1em}
	\label{failure}
\end{figure}

There are two major limitations of AGE. The first one comes from the sampling during inference. This sampling is based on the statistics of the training set rather than adaptive to the input images. 
As shown in Figure~\ref{failure}, if the objects are of an irregular posture (\eg dogs in sideways), the generated images are more likely to be crashed. Moreover, some attributes are not category-irrelevant for all categories, but they will be learned in $\bm{A}_f$. For example, the number of petals is category-irrelevant for most flowers, but it may be identical for some specific categories. Although the generated images are realistic, the category has been changed. 

Moreover, the performance of AGE largely relies on the pretrained styleGAN and the inversion method. If the category-relevant attributes of input image can not be well embedded, the editing will also fail. 
In future, we will try to factorize both category-relevant and irrelevant attributes with better disentanglement, and train the GAN inversion and attribute factorization end to end.

\section{Conclusion}
In this work, we present a new editing-based perspective for few-shot image generation. In order to achieve category-irrelevant editing on unseen classes, we propose a new method, Attribute Group Editing (AGE), which can identify category-relevant and irrelevant semantics learned by GANs without explicit supervision. Extensive experiments demonstrate the diversity and stability of our method on both few-shot image generation and controllable image manipulation on category-irrelevant attributes.

\section{Acknowledgement.} This work was supported in part by the National Key R\&D Program of China under Grant 2018AAA0102000, in part by National Natural Science Foundation of China: 62022083, U21B2038, 61931008 and 61836002, in part by the Fundamental Research Funds for the Central Universities.

\clearpage

\clearpage
\section*{\LARGE Supplementary Material}
\appendix
This supplementary document is organized as follows:
\begin{itemize}
    \item \cref{vgg} provides the ablation study for the number of test categories in VGGFaces~\cite{vggfaces} (Section 4.4).
    \item \cref{gaussian} provides the demonstration of the assumption of Gaussian distribution of $\mathcal{W^+}$ space.
    \item \cref{sefa} provides visualizations of the interpretable semantics discovered by unsupervised image manipulation method SeFa~\cite{sefa}. It is not able to handle multi-class image generation nor distinguish category-relevant and category-irrelevant attributes like AGE.
    \item \cref{sample} provides visualizations of the ablation ``Sample Train" (Section 4.3).
    \item \cref{good} provides additional visualizations of one-shot image generalization from AGE (Section 4.5).
    \item \cref{failurecase} provides additional visualizations and analysis of failure cases from AGE (Section 4.7).
    \item \cref{svd1} provides additional visualization of disentangled attribute editing directions after SVD (Section 4.6).
\end{itemize}


\section{Quantitative Results on VGG Faces Test Split \label{vgg}}
In the experiment, another interesting phenomenon is that FID and LPIPS is highly correlated with the number of categories in the test set.
For fair comparison, we test our AGE model on VGGFaces~\cite{vggfaces} with different numbers of categories in the test split. The quantitative test results are shown in Table~\ref{tab:500}. We can find that the more categories in the test split, the lower FID score and higher LPIPS score the model will get. 
This is conducive to a more comprehensive evaluation of the model. Our AGE model achieves a better quantitative result with FID $34.86$ and LPIPS $0.3294$ when there are 572 categories in the test split.

\begin{table}[h]
\centering
\setlength{\tabcolsep}{7mm}
\caption{Ablations of different numbers of categories in the test split on VGG Faces.}
\begin{tabular}{@{}cccc@{}}
\toprule
\multirow{2}{*}{\begin{tabular}[c]{@{}l@{}} \# Categories \end{tabular}} & \multicolumn{2}{c}{VGG Faces}    \\
                                                                                 & FID(↓)         & LPIPS(↑)        \\ \hline
2                                                                                & 78.83         & 0.2974          \\
50                                                                               & 41.07          & 0.3189          \\
200                                                                              & 36.09          & 0.3212          \\
572                                                                              & \textbf{34.86} & \textbf{0.3294} \\ \bottomrule
\end{tabular}
\label{tab:500}
\end{table}

\section{Demonstration of the Assumption of Gaussian Distribution. \label{gaussian}}

\begin{figure}[t]
\centering
     \includegraphics[width=\linewidth]{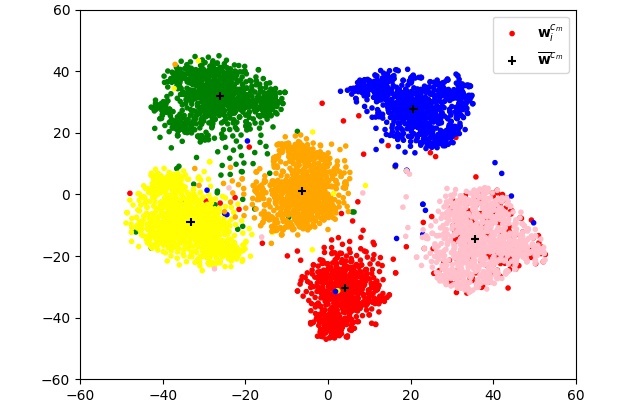}
     \vspace{-2em}
    \caption{Visualization of $\mathcal{W^+}$ space after TSNE.}
    \label{fig:tsne}
\end{figure}

We make an assumption that the distribution of the samples in $\mathcal{W^+}$ space obeys Gaussian distribution in Eq. 17.
This assumption is from StyleGAN that different images can be generated from a center image with linearly interpolation along different directions in the embedding space.
In Figure~\ref{fig:tsne}, we further illustrate the latent embeddings of samples from 6 different categories after TSNE. The distribution of different categories does roughly follow Gaussian distribution.

\section{Comparison with Unsupervised Image Manipulation Methods \label{sefa}}

\begin{figure*}[t]
\centering
     \includegraphics[width=\linewidth]{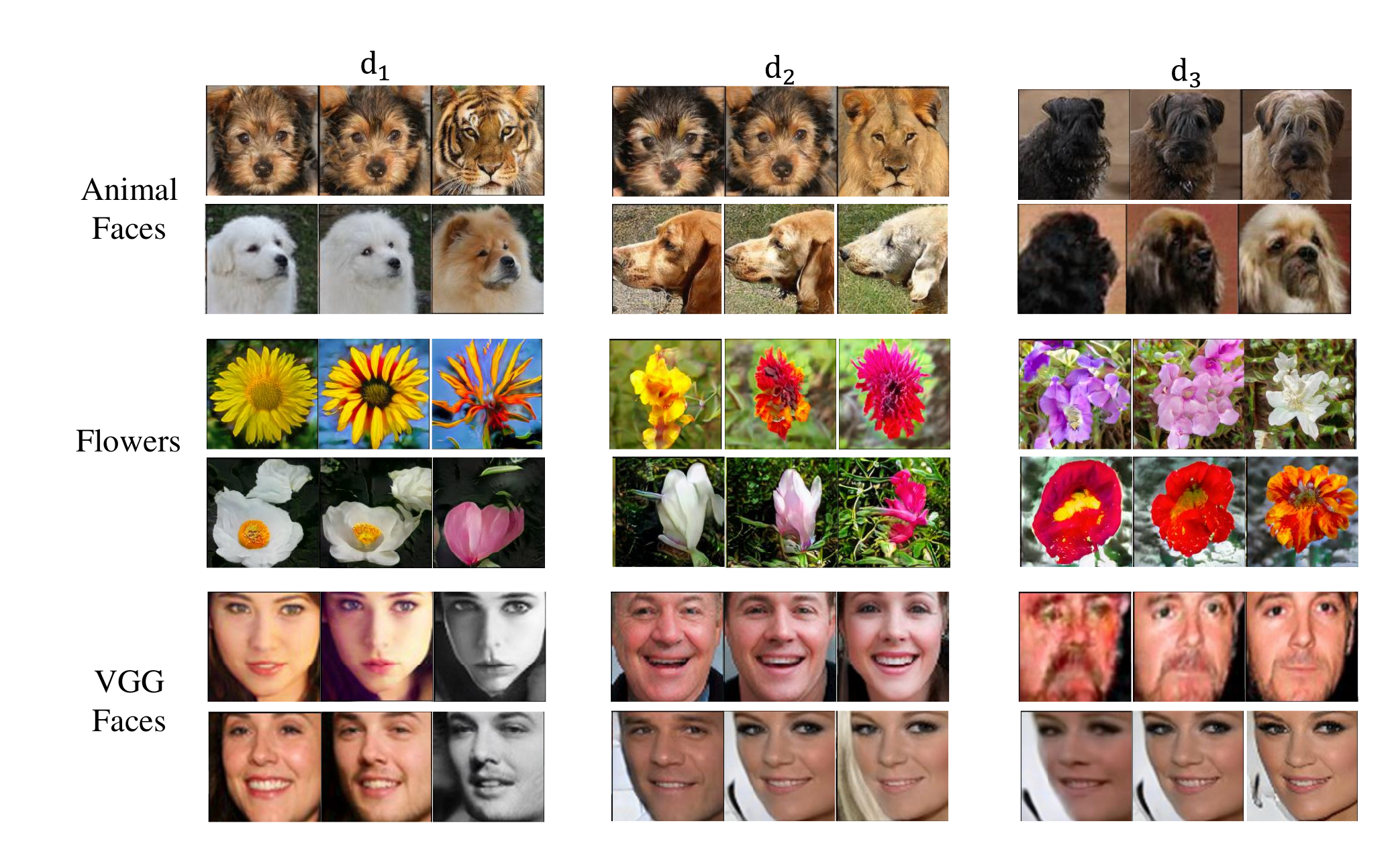}
     \vspace{-2em}
    \caption{Visualizations of interpretable directions discovered by SeFa. The left and the right images are edited from the middle one.
    Moving the latent vectors along the discovered directions apparently changes the categories of the images.}
    \label{fig:sefa}
\end{figure*}

The core of the editing-based few-shot image generation is to identify the category-relevant and category-irrelevant attributes in the latent space without explicit supervision.
Similar to AGE, unsupervised image manipulation methods~\cite{sefa,RW11-Hessian,jacobian} also study the semantic factorization of a pre-trained GAN. 
However, they only focus on single-category image generation that does not care about the categorical information.
In order to verify if they can distinguish the category-irrelevant directions for few-shot image generation, we conduct the recent proposed method SeFa~\cite{sefa} on three multi-class image generation datasets.
SeFa performs a closed-form factorization on the latent semantics according to the weights of the generator, which is one of the best unsupervised attribute factorization and manipulation methods.

Figure~\ref{fig:sefa} shows the first three directions discovered by SeFa. In complicating datasets Animal Faces~\cite{animalfaces} and Flowers~\cite{flowers}, the category-irrelevant attributes and category-relevant attributes are all entangled. The interpretable semantics are hard to distinguish. Editing along a single direction changes multiple attributes and results in an image of a completely different category (\eg from a dog to tiger, the shape of the petals, etc.). In VGGFaces~\cite{vggfaces}, despite achieving better disentanglement, the top important semantics discovered by SeFa are almost category-relevant including the sex and the shape of the face. In contrast, AGE can factorize the category-irrelevant attributes from the category-relevant attributes, which is the most important for few-shot image generation.



\begin{figure*}[ht]
    \centering
    \includegraphics[width=0.95\linewidth]{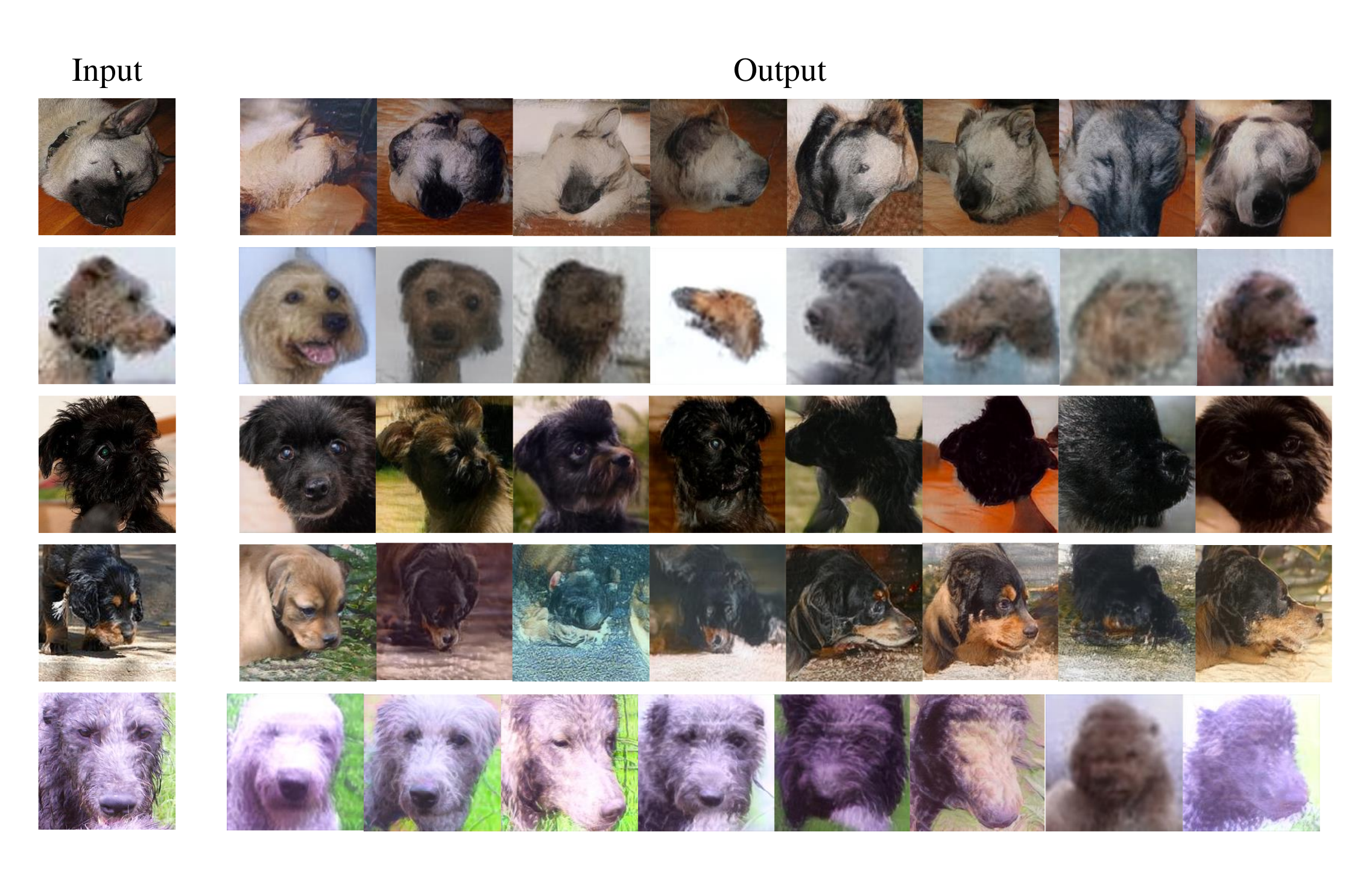}
   \vspace{-0.5em}
    \caption{Images edited by random $\Delta\mathbf{w}$ sampled from seen categories.}
    \label{fig:sampletrain}
\end{figure*}

\section{Images Generated from ``Sample Train" \label{sample}}
In Section 4.3, we provide the ablation ``Sample Train" that randomly samples $\Delta\mathbf{w}$ of seen categories from the train set and directly use it to edit the unseen categories. As shown in Table~2 in the main paper, it degrades a lot on FID compared with AGE.

In this section, we provide samples generated from ``Sample Train" in Figure~\ref{fig:sampletrain}. As shown in the generated samples, directly using the sampled $\Delta\mathbf{w}$ to edit the input images is very unstable. Although some high-quality images can be generated, most images are crashed or change category.
This also further proves the necessity of the attribute factorization of AGE.


\begin{figure*}[ht]
    \centering
    \includegraphics[width=0.9\linewidth]{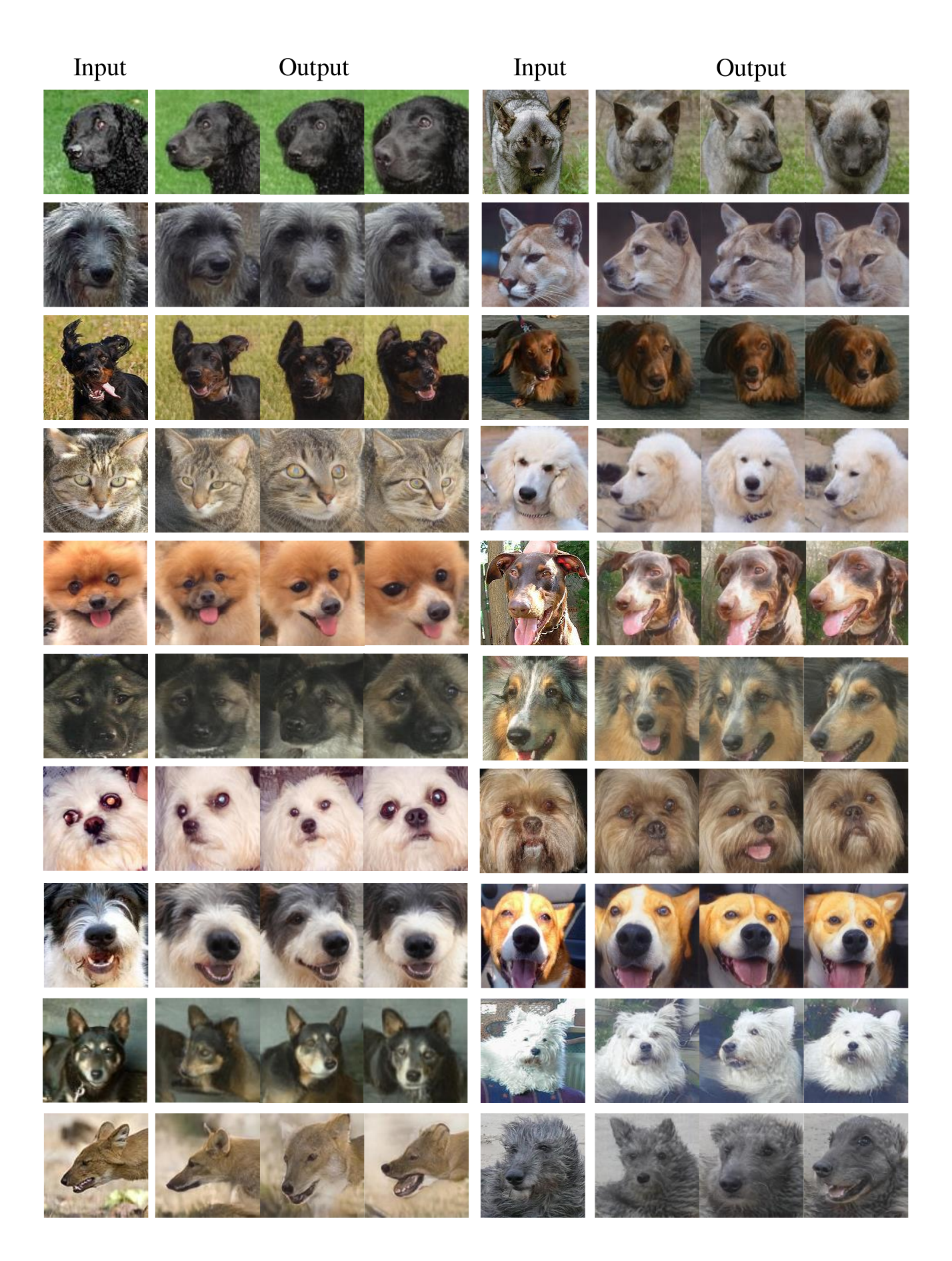}
   \vspace{-0.5em}
    \caption{One-shot image generation by AGE on Animal Faces.}
    \label{fig:good1}
\end{figure*}

\begin{figure*}[ht]
    \centering
    \includegraphics[width=0.9\linewidth]{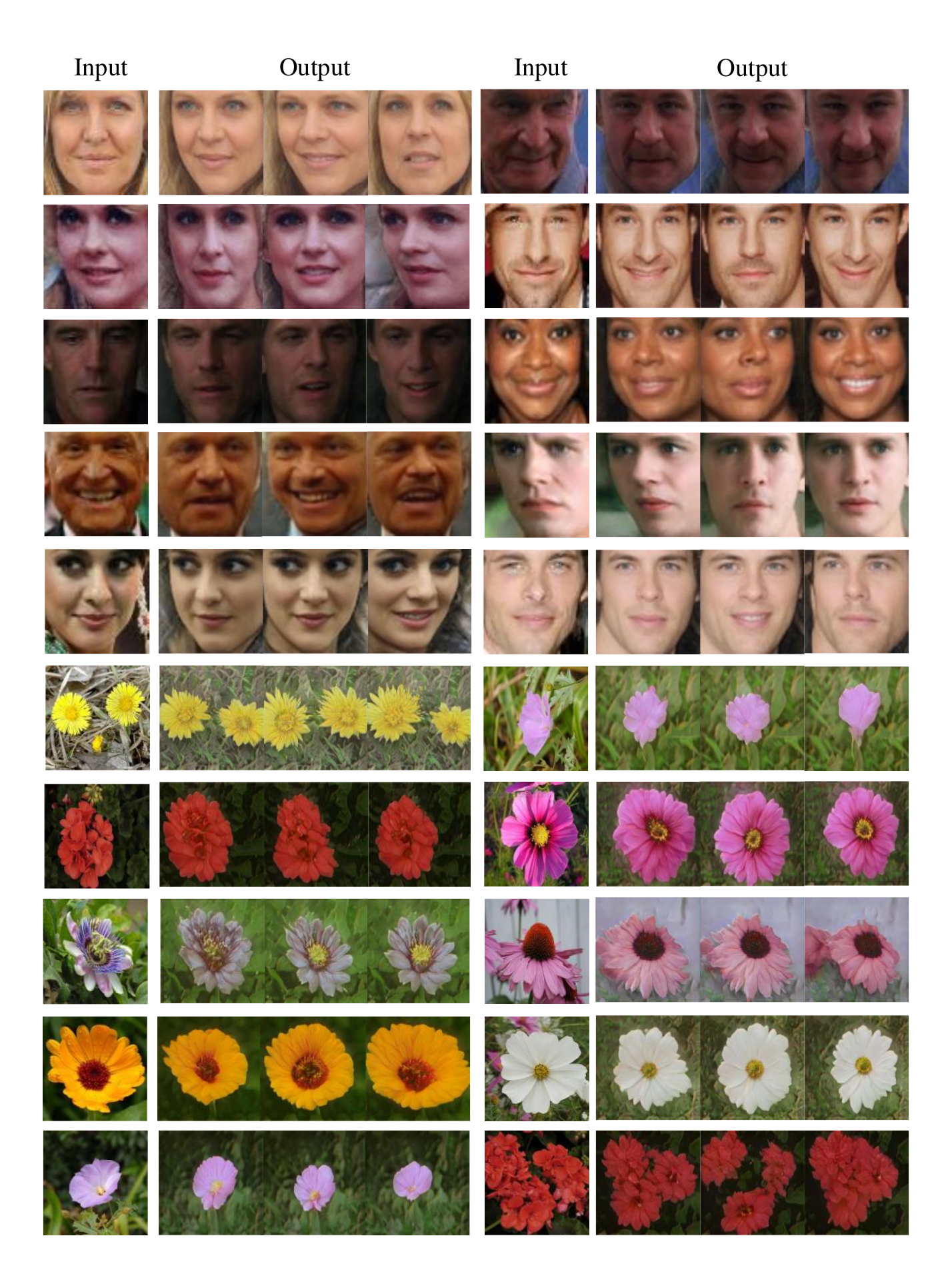}
   \vspace{-0.5em}
    \caption{One-shot image generation by AGE on VGGFaces and Flowers.}
    \label{fig:good2}
\end{figure*}
\section{Additional Visualizations for AGE \label{good}}

We provide more samples generated by AGE in Figure~\ref{fig:good1} and Figure~\ref{fig:good2}.


\begin{figure*}[t]
	\begin{center}
		\subfloat[Inversion Failure]{
			\label{subfig:if}
			\includegraphics[width=\linewidth]{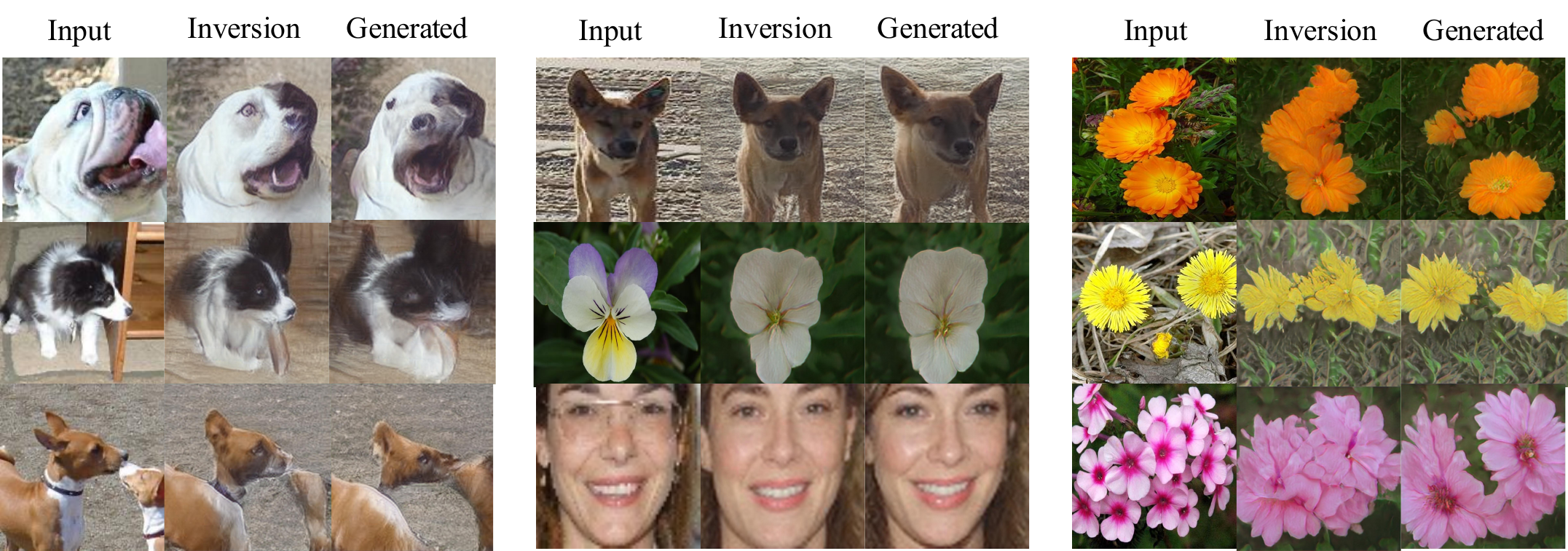}
		}\\
        \vspace{1em}
		\subfloat[Category Change]{
			\label{subfig:cc}
			\includegraphics[width=\linewidth]{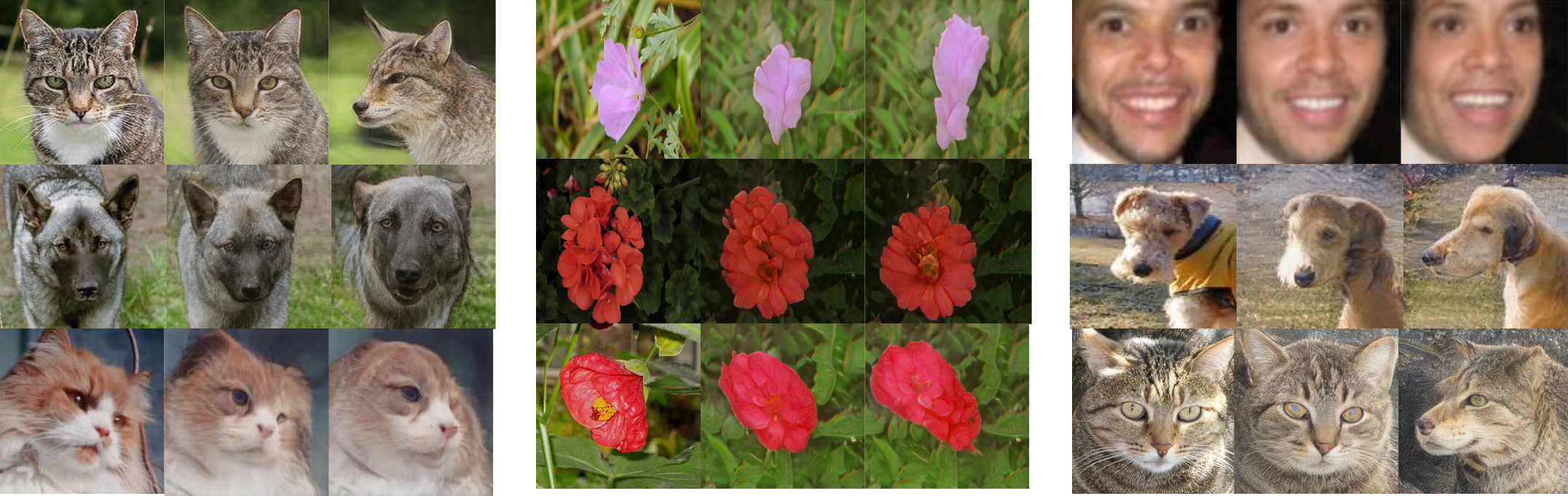}
		}
		\\	
		\vspace{1em}
		\subfloat[Editing Failure]{
			\label{subfig:ef}
			\includegraphics[width=\linewidth]{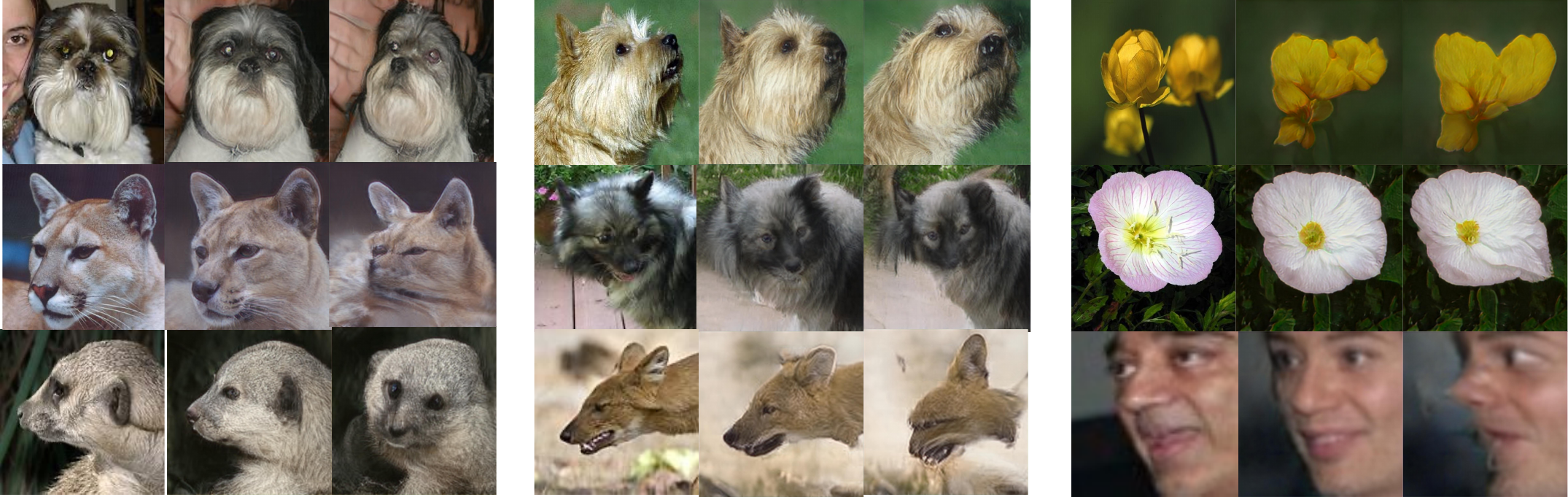}
		}
		
	\end{center}

	\caption{Failure Cases from AGE.
	}
	\label{fig:failure}

\end{figure*}

\section{Failure Case Analysis for AGE \label{failurecase}}

We provide more failure cases generated by AGE in Figure~\ref{fig:failure}.
Failure cases can be divided into three classes: inversion failure, category change, and editing failure. 
Most crashed cases of AGE are caused by the failure of GAN inversion as shown in the Figure~\ref{subfig:if}.
Our editing starts from the latent representation of GAN inversion. Therefore, if the inversion representations cannot reconstruct the input images, both of the attribute factorization and manipulation will fail.
GAN inversion is not stable when there aren't enough samples.
In Flowers~\cite{flowers}, many important category-relevant attributes are lost after inversion. In VGGFaces~\cite{vggfaces}, all glasses are missing after inversion, therefore, this attribute is completely ignored during training.

Second, some editing from AGE may cause category change. This is because some category-irrelevant attributes learned by AGE are not shared among all categories (\eg the number of petals and the shape of cats' face). This situation is more common in the Flowers~\cite{flowers} dataset since the intra-category variations of different kinds of flowers are very distinct. 

Third, since the sampling process of sparse representation is based on the statistics of the whole training set, the editing generated from AGE may lead to crashes in the images when encountering extreme cases.
We hope our new editing perspective can inspire further researches towards better attribute disentanglement free from pre-trained GAN inversion methods.

\begin{figure*}[ht]
    \centering
    \includegraphics[width=0.9\linewidth]{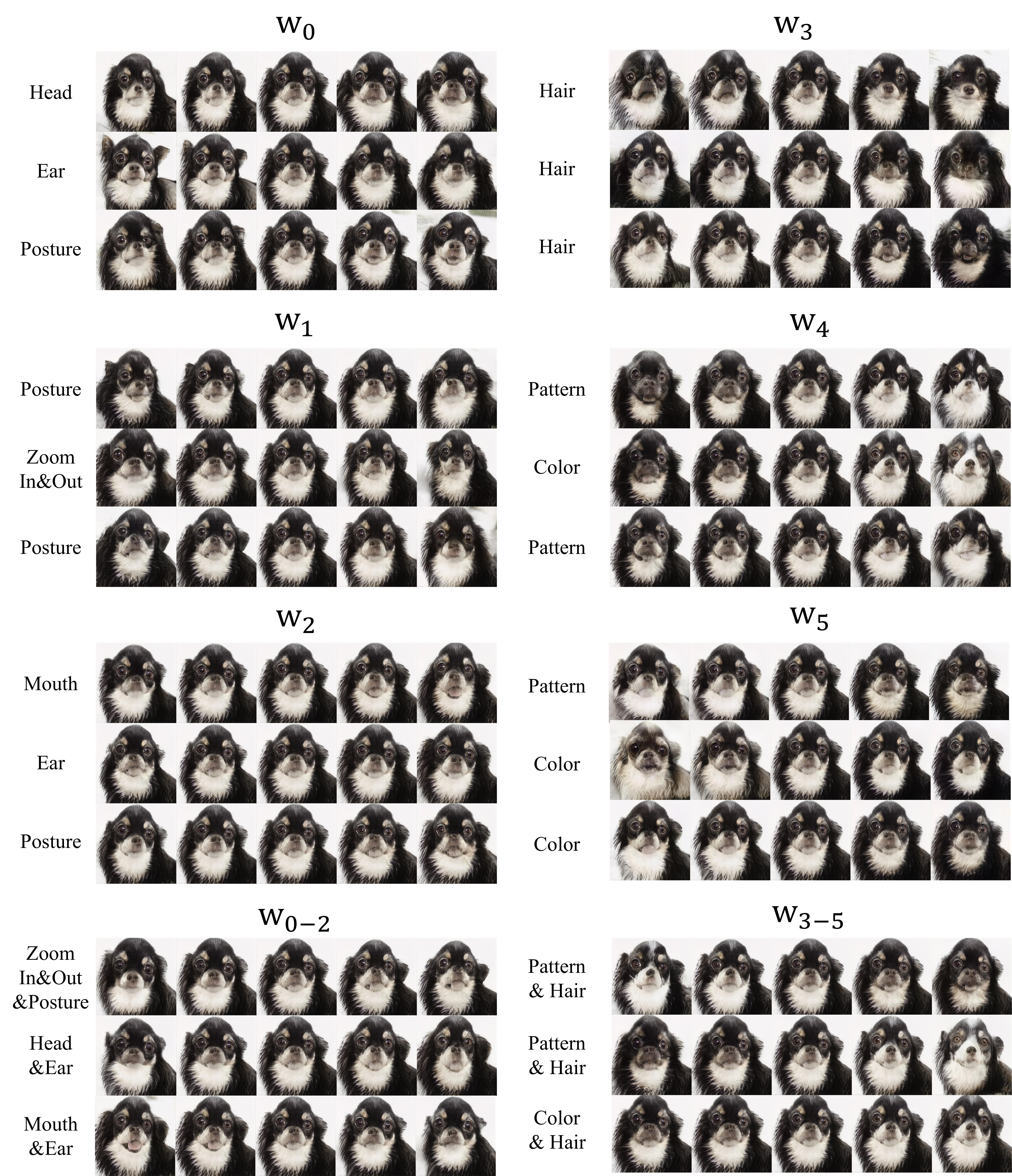}
   \vspace{-0.5em}
    \caption{Visualizations of disentangled attribute editing directions in different layers/groups learned by AGE.}
    \label{fig:svd1}
\end{figure*}

\section{Additional Disentangled Attribute Editing Directions \label{svd1}}

We provide more visualizations of disentangled attribute editing directions in different layers/groups learned by AGE in Figure~\ref{fig:svd1}. These images are edited along the directions factorized with SVD. The details have been provided in Section 4.6 in the main paper.

As shown in Figure~\ref{fig:svd1}, different directions in different layers control different attributes. Editing along certain directions can roughly change one specific attribute continuously. The lower layers like $\mathbf{w}_0$, $\mathbf{w}_1$, $\mathbf{w}_2$ mainly control structure attributes like posture, ear, eye,  and head. The higher layers like $\mathbf{w}_3$, $\mathbf{w}_4$, $\mathbf{w}_5$ mainly control surface attributes like hair and color. This is in line with the rules of most GANs' latent spaces.



\end{document}